# AxiomOcean: Forecasting the Three-Dimensional Structure of the Upper Ocean

Sensen Wu[1,†], Yifan Chen[1,†], Guantao Pu[1,†], Xiaoyao Sun[1,2], Yijun Chen[1], Jin Qi[1,2], Ming Kong[1], Keyi Yang[1], Lichen Xu[1], Wenguan Wang[3], Xiaofeng Li[4], Zhenhong Du[1,2,*]

[1] School of Earth Sciences & Zhejiang Key Laboratory of Geographic Information Science, Zhejiang University, Hangzhou, China

[2] State Key Laboratory of Ocean Sensing, Zhejiang University, Hangzhou, China

[3] The State Key Lab of Brain-Machine Intelligence, Zhejiang University, Hangzhou, China

[4] Key Laboratory of Ocean Circulation and Waves, Institute of Oceanology, Chinese Academy of Sciences, Qingdao, China.

## Abstract

Short-term ocean forecast skill depends strongly on the three-dimensional ocean structure of the upper ocean, which governs stratification, subsurface heat storage, and the response of the ocean to atmospheric forcing. However, AI ocean forecasting models often fail to preserve this vertical structure, resulting in over-smoothed subsurface features and weak physical consistency under strong forcing. Here, we present AxiomOcean, a global AI ocean forecasting model that explicitly represents vertical hierarchy and cross-layer dependence within the water column. By combining a fully three-dimensional encoder-backbone-decoder architecture with surface atmospheric forcing, AxiomOcean jointly predicts upper-ocean temperature, salinity, and three-dimensional currents at global 1/12° resolution down to 643 m depth. In 10-day forecasts, AxiomOcean outperforms an advanced AI comparison model across variables and lead times, reducing day-1 RMSE by approximately 20 to 35% while maintaining higher anomaly correlation. The gain is not achieved through excessive smoothing: AxiomOcean better preserves eddy kinetic energy, temperature and salinity variance. Its advantage also extends through the water column and remains evident across the equatorial Pacific, Kuroshio Extension, and Southern Ocean, yielding a more realistic reconstruction of upper-ocean heat content. These results show that explicitly preserving upper-ocean three-dimensional structure can improve both forecast accuracy and physical fidelity in AI ocean prediction.

† These authors contributed equally to this work.
* Corresponding author: Zhenhong Du (duzhenhong@zju.edu.cn).

# Introduction

Daily-to-weekly ocean variability is controlled by coupled three-dimensional dynamical and thermodynamical processes within the upper ocean, including stratification, mixed-layer adjustment, and subsurface heat redistribution [1]. These processes shape the exchange of heat, momentum, and salt across the water column and strongly influence the ocean response to atmospheric forcing [2,3]. Subsurface thermal structure and thermocline variability also provide memory that regulates the evolution of marine heat extremes, air-sea interaction, and tropical-cyclone feedbacks [4-10]. For short-term ocean forecasting, realistic representation of upper-ocean vertical structure is therefore a prerequisite for physical credibility.

Global ocean forecasting has long relied on numerical ocean circulation models that integrate the governing equations under prescribed forcing and boundary conditions [11,12]. These models explicitly represent three-dimensional physical processes and remain indispensable for process understanding. However, achieving eddy-resolving resolution, sufficiently rich vertical structure, and frequent global forecast updates remains computationally expensive [13-15]. Forecast quality is also sensitive to initial conditions, boundary forcing, and model structural error. As a result, even advanced numerical systems can struggle to maintain realistic upper-ocean thermal structure and subsurface adjustment in short-range forecasting.

Recent progress in artificial intelligence has created a new pathway for geophysical prediction. In weather forecasting, data-driven models now achieve remarkable skill at far lower computational cost than traditional numerical systems [16-20]. This success has rapidly motivated parallel developments in ocean prediction, including regional neural forecasting systems, digital twins, and emerging global AI ocean models [21-25]. Yet the ocean poses a distinct challenge: short-term forecast skill depends strongly on the integrity of a stratified water column. Many existing AI ocean models flatten the three-dimensional ocean state into collections of parallel two-dimensional channels. That design weakens the layer-to-layer relationships that organize density structure, thermocline geometry, mixed-layer adjustment, and vertical transport. It can yield competitive pointwise scores while still producing over-smoothed profiles, degraded cross-variable consistency, and weakened ocean response under strong atmospheric forcing [26-29].

Here we introduce AxiomOcean, a deep-learning framework for forecasting upper-ocean vertical three-dimensional structure, which utilizes a fully three-dimensional encoder-backbone-decoder architecture to retain layered organization and cross-layer dependence within the water column. Incorporating eight atmospheric forcing variables as external boundary information, the system performs global 1/12° forecasting of temperature, salinity, and three-dimensional currents over the upper 643 m of the ocean. In this study, we demonstrate that this explicit structural design improves upon deterministic forecast skill baselines, reducing root mean square error while increasing the anomaly correlation coefficient over 10-day forecasts. Furthermore, we show that AxiomOcean avoids artificial gradient smoothing, preserving vital mesoscale spatial structures like eddy kinetic energy and spatial variance. By testing across distinct dynamical regimes, we reveal that the model consistently preserves thermocline sharpness and depth-dependent adjustment mechanisms. Ultimately, these combined improvements yield a highly accurate representation of upper-ocean heat content, confirming the physical consistency and predictive viability of the model under atmospheric forcing.

# Results

We evaluated AxiomOcean in 10-day global forecasts initialized from the preceding two daily ocean states together with atmospheric forcing, using GLORYS-based reanalysis fields for verification [30,31].

## Advanced multivariate forecasting performance of AxiomOcean

At the global scale, AxiomOcean consistently outperforms an advanced AI comparison model [25] in upper-ocean temperature, salinity, and current prediction.

As shown in Figure 1, across lead times from day 1 to day 10, the model yields lower root mean square error (RMSE) and higher anomaly correlation coefficient (ACC) than the comparison model. On day 1, RMSE decreases by approximately 20% to 35%, and the advantage remains evident throughout the 10-day forecast window.

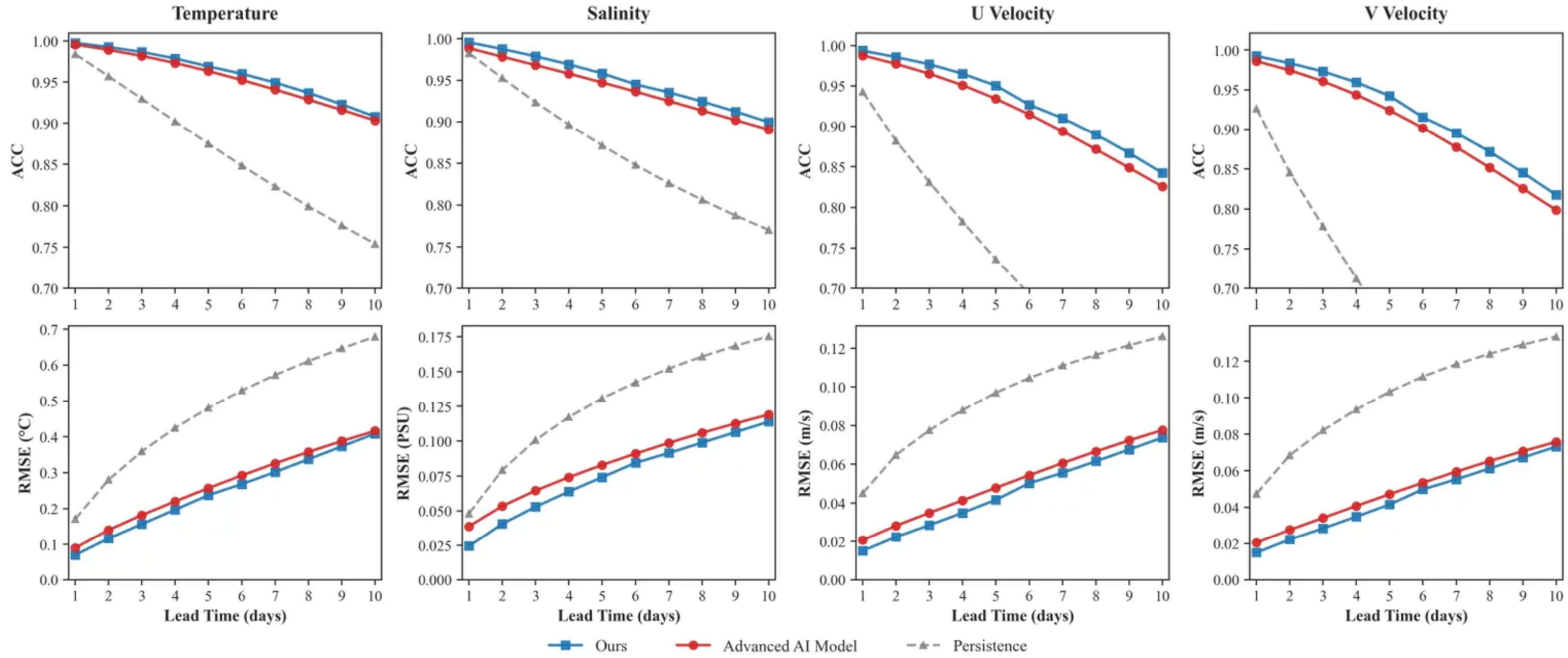

**Fig.1 Global deterministic forecasting capability of AxiomOcean for upper-ocean temperature, salinity, and currents.** A comparison of AxiomOcean's multi-day deterministic forecasts with the advanced AI comparison model. This figure uses root mean square error (RMSE) and anomaly correlation coefficient (ACC) to illustrate the forecasting performance of upper-ocean temperature, salinity, and horizontal current variables over lead times from day 1 to day 10.

Lower RMSE alone, however, does not guarantee that a high-resolution ocean forecast retains dynamically relevant structure [32,33]. Mesoscale variability, frontal sharpness, and spatial heterogeneity govern energy distribution and tracer transport and are therefore integral to physically meaningful prediction [27,34]. Models optimized primarily against mean-squared error often reduce error by damping gradients and suppressing variability, producing overly smooth forecast fields [35]. The key question is thus whether forecast gains can be achieved without erasing the spatial signatures of ocean dynamics.

To address this issue, we diagnosed volumetric eddy kinetic energy (EKE), temperature variance, and salinity variance in the forecast fields. These metrics constrain the spatial expression of mesoscale energy and upper-ocean heterogeneity [36,37]. From the results presented in Table 1, it can be observed that AxiomOcean remains closer to the reanalysis in all three measures, whereas the comparison model shows systematic variance loss and EKE attenuation. The improvement therefore reflects better preservation of dynamically relevant structure rather than simple variance damping.

**Table 1 Diagnostic results of mesoscale phenomena and spatial variability in forecast results**

| | **Eddy Kinetic Energy (EKE)** | **Temperature Variance** | **Salinity Variance** |
|---|---|---|---|
| **Truth** | 2.26E+15 | 8.50E+16 | 3.4E+15 |
| **AxiomOcean** | 2.2E+15 | 8.22E+16 | 3.3E+15 |
| **Comparison model** | 2.1E+15 | 7.96E+16 | 3.27E+15 |

We also examined local structure through the distribution of sea surface temperature (SST) gradients. Gradient statistics provide a direct test of whether fronts and fine-scale sharpness are retained or artificially diffused. As shown in Figure 2, AxiomOcean better reproduces the observed gradient distribution overall, whereas the comparison model exhibits clear gradient attenuation. Together, these results show that explicit vertical structural modeling improves both pointwise forecast skill and the preservation of multiscale spatial structure.

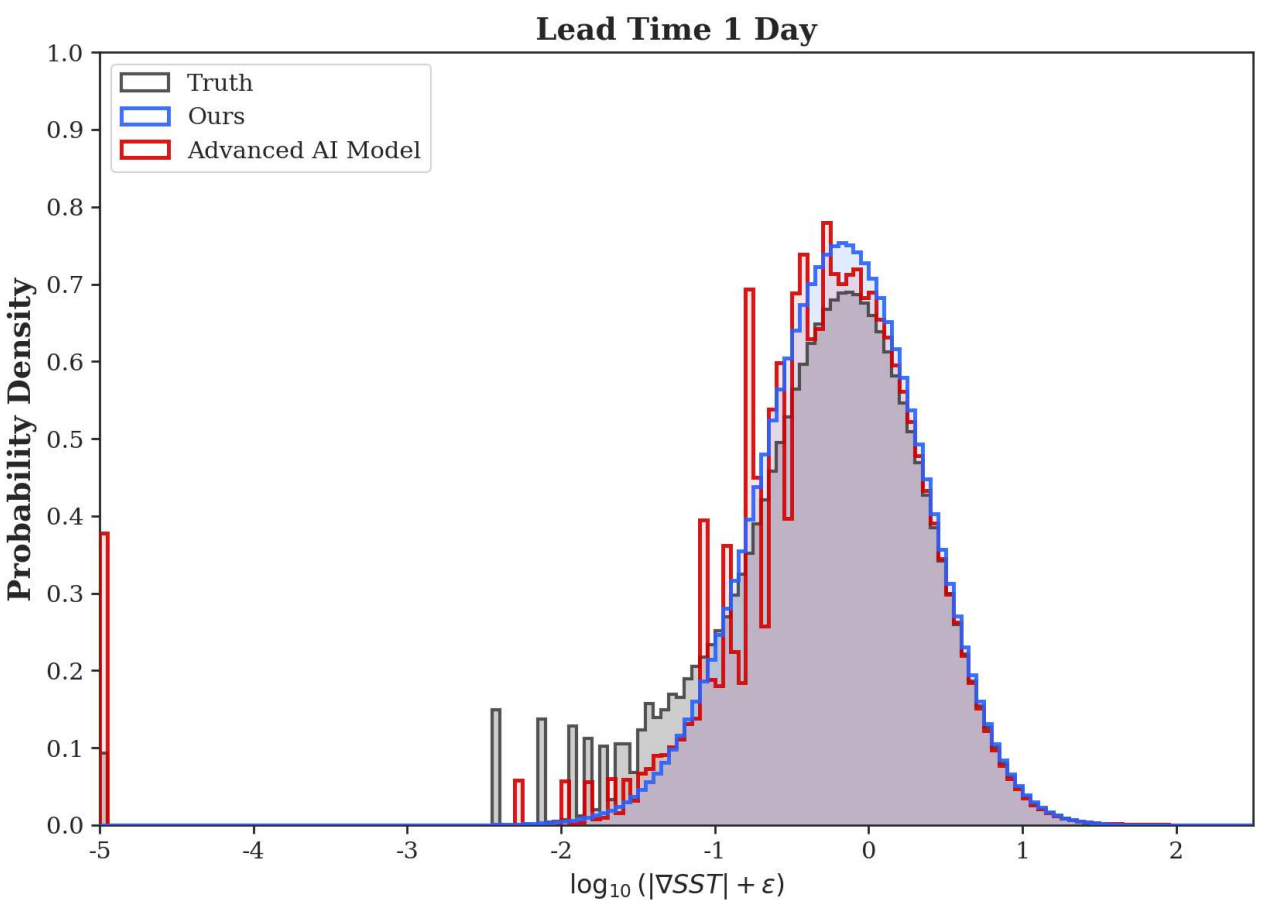

**Fig.2 AxiomOcean's ability to preserve small-scale spatial structures.** Statistical evaluation of sea surface temperature (SST) gradient structures is used to diagnose whether forecasts retain frontal and spatial sharpness, rather than reducing errors through over-smoothing.

# Forecast gains extend through the upper-ocean vertical structure

Because the ocean is strongly stratified, predictive skill in the vertical dimension is essential for physical reliability [38,39]. Figure 3 shows that AxiomOcean yields lower depth-dependent RMSE for temperature, salinity, and currents through the upper 643 m, and that the separation from the comparison model becomes especially clear away from the immediate surface layer. This depth-dependent behavior indicates that the model improvement is not limited to near-surface signals directly constrained by atmospheric forcing, but extends into the upper-ocean structure that governs delayed adjustment and vertical redistribution.

This distinction matters because subsurface evolution is less directly constrained by instantaneous forcing than the surface ocean and depends more strongly on the vertical arrangement of heat, salt, and momentum. A forecast system can therefore achieve competitive near-surface scores while still failing to recover thermocline displacement, subsurface storage, or vertically coherent current adjustment. The profile-based results suggest that AxiomOcean better captures these deeper structural components of forecast evolution.

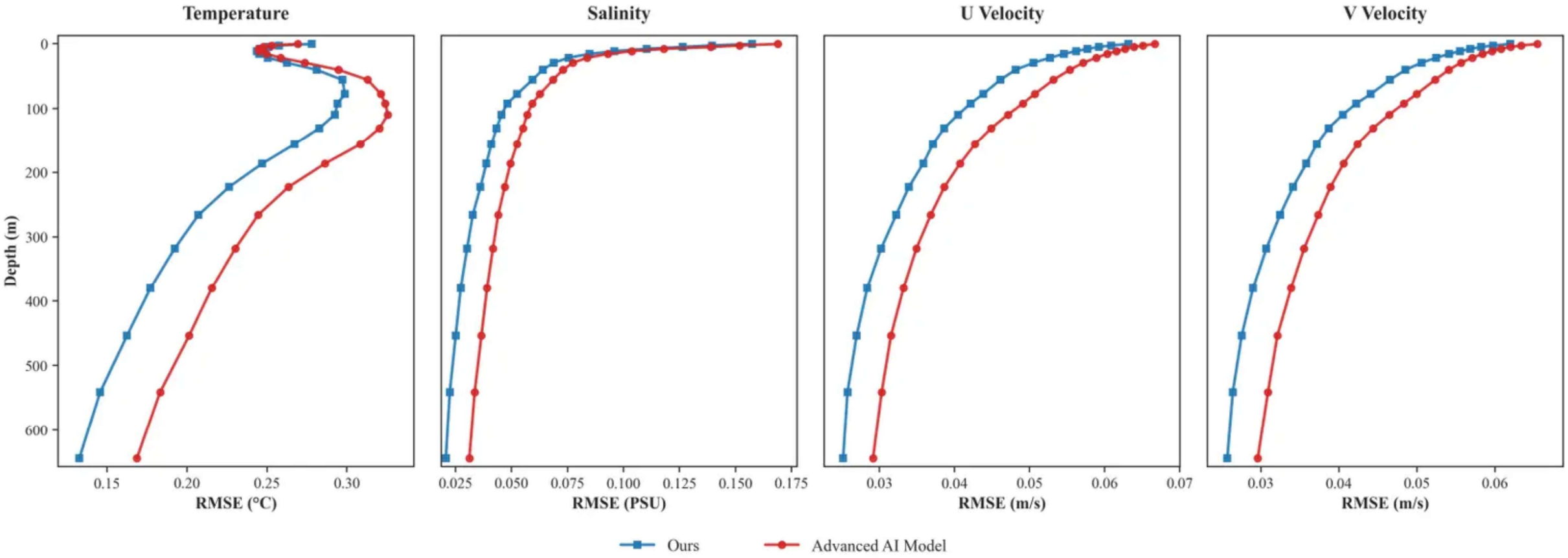

**Fig.3 AxiomOcean's predictive capabilities for various elements on vertical profiles.** AxiomOcean, along with the comparison model, assesses the prediction error (RMSE) of ocean temperature, salinity, and current fields at various depths.

The likely source of this gain is the model's explicit three-dimensional treatment of the water column. Architectures that flatten depth into parallel 2D channels can still optimize global RMSE, but they tend to weaken cross-layer relationships that organize thermocline geometry, mixing transitions, and vertically coherent current structure. In contrast, AxiomOcean preserves layered organization during encoding and evolution, which is consistent with the stronger upper-ocean performance seen here. The depth-dependent forecast improvements are therefore not just an additional benchmark; they provide direct evidence that the architectural choice matters for representing upper-ocean physical memory.

## Structural advantages persist across contrasting dynamical regimes

To test whether the vertical-skill advantage generalizes across different dynamical settings, we examined temperature-section behavior in three representative regions: the equatorial Pacific (0°, 140°E–100°W), the Kuroshio Extension (40°N, 140°E–180°), and the Southern Ocean (55°S, 60°E–150°E), as shown in Figure 4.

These regions place distinct demands on a forecast system. The equatorial Pacific emphasizes thermocline adjustment under wind and wave forcing [40]; the Kuroshio Extension tests whether the model can preserve sharp gradients and sloping isotherms in a strongly baroclinic frontal regime [41,42]; and the Southern Ocean probes performance under deep wind-driven mixing and vertically integrated redistribution [43,44].

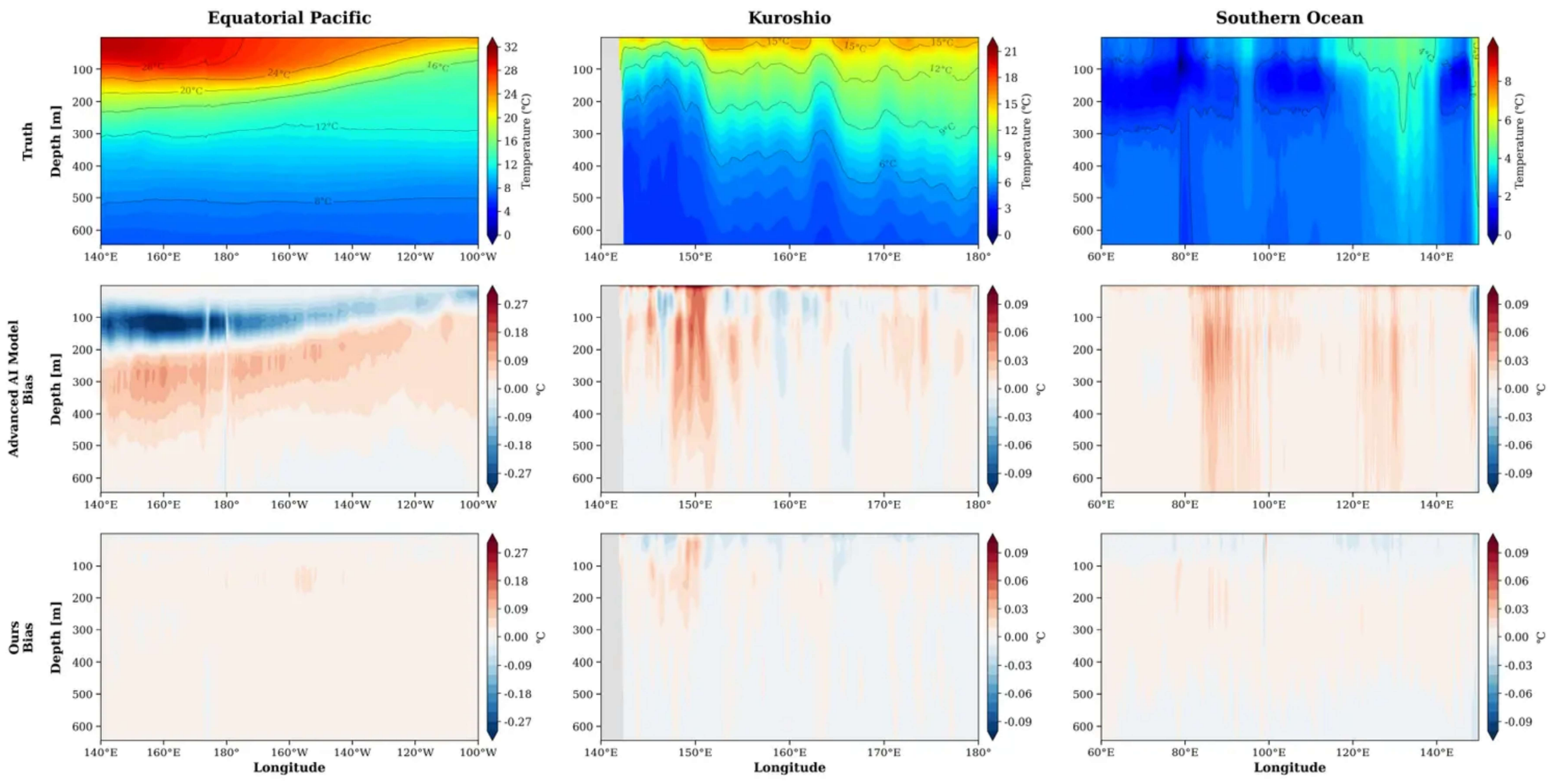

**Fig.4 Temperature profile errors in three typical upper-ocean regions with different dynamic modes.** Comparison of vertical temperature structure errors in three representative ocean regions used to diagnose and improve physical causes.

In the equatorial Pacific, AxiomOcean reduces systematic temperature errors near the thermocline, where the comparison model shows a stronger tendency toward structural distortion and weakened stratification. This result is important because thermocline depth and sharpness govern upper-ocean heat storage and the transmission of wind-forced signals. Better recovery of the thermocline indicates that the model more faithfully represents vertically coupled adjustment. In the Kuroshio Extension, AxiomOcean more effectively preserves the slope and continuity of the isotherm field, while the comparison model shows stronger structural diffusion. This is precisely the kind of regime in which spatial sharpness and vertical coherence must be retained simultaneously. In the Southern Ocean, where strong winds and deep mixing broaden the depth range of active adjustment, AxiomOcean suppresses water-column-scale bias more effectively than the comparison model.

The improvement therefore changes form across regions, but not randomly: the model sharpens stratified interfaces where thermocline structure dominates, better preserves frontal geometry where baroclinicity is central, and better constrains vertically integrated redistribution where mixing is strong [45,46]. This consistency across regimes strengthens the mechanistic interpretation of the global benchmark gains.

## Physical consistency of the upper-ocean structure from forecasting results

The preceding sections show that AxiomOcean improves multivariate forecast skill, better preserves spatial variability, and maintains stronger vertical structure across contrasting regimes. The remaining question is whether these improvements translate into a more realistic integrated state of the upper ocean rather than remaining confined to variable-by-variable error reduction. To address this, we examined upper-ocean heat content (OHC) integrated over 0 to 100 m, which provides a compact measure of the thermal reservoir involved in short-term air-sea interaction.

This depth range captures much of the mixed-layer signal together with the upper thermocline response, making OHC an especially useful diagnostic of whether improvements in predicted temperature fields remain physically coherent after vertical integration. In this sense, OHC is a more demanding test than pointwise temperature error alone: a model must not only reduce local error, but do so in a way that preserves the cumulative thermal structure of the upper ocean.

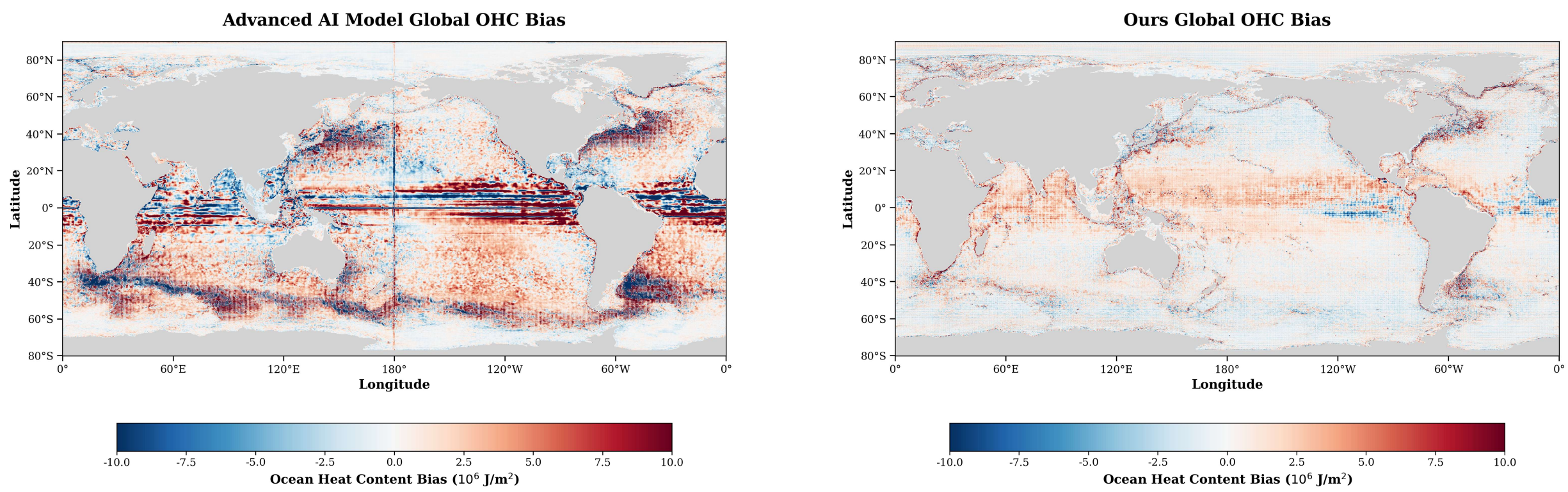

**Fig.5 Spatial error distribution and temporal evolution of upper-ocean heat content in forecasts.** Upper-ocean heat content (OHC; 0–100 m) assessment as a comprehensive measure of thermodynamic structure related to air-sea interaction. This figure compares the spatial distribution and temporal evolution of upper-ocean heat content in the AxiomOcean model with the comparison model and contrasts it with the reanalysis true values.

As shown in Figure 5, AxiomOcean reconstructs the spatial distribution of OHC more accurately than the comparison model. The bias patterns are generally weaker and less spatially coherent, indicating that the model better represents the large-scale distribution of upper-ocean thermal storage. OHC links the local structural improvements shown in Figures 3 and 4 to a physically consequential state variable. A more realistic OHC field therefore implies a more realistic representation of the upper ocean's capacity to store and redistribute heat under short-term forcing [47]. This diagnostic shows that the model improvement extends beyond conventional forecast scores to the physically emergent organization of the upper ocean.

# Discussion

We have presented AxiomOcean, a global AI ocean forecasting system that explicitly represents the three-dimensional vertical structure of the upper ocean. Unlike approaches that treat depth levels mainly as parallel channels, AxiomOcean encodes the water column as a structured three-dimensional system in which layered organization and cross-layer dependence are maintained throughout the forecast process. This design improves not only forecast skill for temperature, salinity, and three-dimensional currents, but also the physical consistency of upper-ocean stratification and heat-content evolution.

The broader significance of these results is conceptual as much as technical. For short-term ocean forecasting, the upper ocean is not simply a collection of near-surface variables; it is a vertically organized memory system that controls how the ocean integrates and redistributes atmospheric forcing. Our findings show that by explicitly modeling this three-dimensional structure, AxiomOcean helps limit the over-smoothing of vital dynamic signatures. Specifically, the model retains critical mesoscale spatial variability, preserving EKE and sea surface temperature gradients rather than damping them to minimize pointwise errors. Furthermore, vertical predictive skill is maintained well below the immediate surface layer, indicating a more faithful representation of subsurface storage and delayed adjustment mechanisms.

Crucially, these structural advantages persist across fundamentally different dynamical regimes. AxiomOcean actively reduces systematic errors in the wind-forced thermocline of the Equatorial Pacific, the baroclinic frontal regime of the Kuroshio Extension, and the deep-mixing environment of the Southern Ocean. Ultimately, these localized and structural improvements culminate in a highly accurate reconstruction of global upper-ocean heat content. This suggests the model accurately captures the ocean's capacity to store and redistribute heat under short-term forcing, effectively linking local structural gains to a physically consequential state variable.

Preserving that physical organization appears necessary for AI forecasts that remain physically credible across lead times and forcing regimes. In this sense, the present results support a shift in AI ocean prediction from statistical approximation of observed fields toward explicit representation of physical structure. Better forecasts of upper-ocean structure should improve our ability to represent air-sea coupling, upper-ocean heat storage, and rapid ocean adjustment under variable atmospheric forcing.

The current system nevertheless has important limitations. First, the forecast domain extends only to 643 m and does not yet include the deeper ocean, where slower but dynamically relevant adjustment can influence upper-ocean memory. Second, the framework is limited to physical variables and does not yet include ecological or biogeochemical tracers such as chlorophyll or nutrients. To address these constraints, future work should extend the vertical domain toward 2000 m or the full water column so that deeper structural memory can be explicitly learned. Additionally, researchers should further refine spatial resolution to better represent strongly nonlinear mixing processes, and expand the training data with ecological observations to support physically and biologically coupled prediction. These steps would move AI ocean forecasting toward a more general three-dimensional Earth-system representation rather than a surface-focused predictive surrogate.

# Materials and Methods

## Data

Ocean state variables were derived from the GLORYS12V1 global ocean reanalysis [13,30]. The dataset has a horizontal resolution of 1/12°, sufficient to represent mesoscale variability and major western boundary current structures. Because the key daily-to-weekly adjustments of temperature, salinity, and velocity in short-term forecasting are concentrated in the upper ocean, we used 23 standard depth levels spanning 0 to 643 m. The predicted variables were potential temperature, salinity, zonal current, and meridional current, all represented as daily means.

External forcing was taken from ERA5 [14,48]. We used eight near-surface atmospheric variables: 10-m zonal wind (u10), 10-m meridional wind (v10), 2-m air temperature (t2m), 2-m dewpoint temperature (d2m), mean sea level pressure (msl), surface net shortwave radiation (ssr), surface downward longwave radiation (strd), and total precipitation rate (mtpr). These variables represent the major atmospheric controls on upper-ocean momentum, heat, and freshwater adjustment. Ocean and atmospheric fields were regridded to a common spatial grid and temporally aligned at daily resolution before model ingestion.

In this study, we utilized daily data spanning 28 years from 1993 to 2020. To rigorously evaluate the predictive and generalizing abilities of the model, we divided the dataset chronologically. The data from the first 26 years (1993-2018) were used for model training, with 2019 serving as the validation set, and 2020 was selected as the independent test set.

## Overview of prediction framework

The key challenge of high-resolution ocean forecasting lies in the fact that the evolution of oceanic states is simultaneously controlled by internal three-dimensional structural adjustments and external atmospheric forcing. For the upper ocean, short-term changes depend not only on the internal memory determined by existing thermohaline stratification and flow velocity structure, but also continuously respond to surface drivers such as wind stress, heat flux, and freshwater flux. Therefore, ocean forecasting should not be simply regarded as autoregressive extrapolation of historical oceanic states, but rather should be expressed as modeling future evolution under the joint constraints of existing oceanic states and real-time atmospheric forcing.

Based on this understanding, we formulate the forecasting problem as learning ocean-state increments. Here,

the ocean state consists of temperature, salinity, and horizontal currents, and the atmospheric forcing consists of the eight near-surface ERA5 variables described above. Specifically, the model takes the ocean state $(X_{t-1}, X_t)$ and the corresponding atmospheric forcing $(F_{t-1}, F_t)$ at two consecutive moments as inputs, and predicts the state change from the current moment to the future forecast moment:

$$X_{t+\Delta t} = X_t + \mathcal{M}(X_{t-1}, X_t, F_{t-1}, F_t) \quad (1)$$

Where $M$ represents the neural network forecast model and $\Delta t$ denotes the target forecast step size. Compared to directly predicting the absolute future state, this incremental expression can weaken the dominant role of the stable background field in training, allowing the model to focus more on real change signals such as advection transport, vertical mixing, stratification adjustment, and mesoscale structure propagation. Inputting data at two consecutive moments further provides constraints on the trend of state evolution, helping to characterize the inertia and temporal continuity of oceanic changes.

# Model architecture

As shown in Figure 6, AxiomOcean uses a fully three-dimensional encoder-backbone-decoder architecture. The ocean encoder compresses the three-dimensional water-column state into a low-dimensional latent representation that preserves layered thermodynamic and dynamical structure. Atmospheric forcing is encoded separately as a two-dimensional surface-driving representation. The two streams are then fused into a unified latent state describing the coupled ocean-internal and atmosphere-forced configuration of the upper ocean.

The spatiotemporal evolution module adopts a hierarchical Swin Transformer structure. This design is appropriate for ocean forecasting because ocean variability spans strongly interacting scales: large-scale circulation provides background constraint, whereas mesoscale eddies, fronts, and local disturbances dominate short-term error growth. By combining local high-resolution information with broader contextual structure, the architecture captures multiscale evolution without discarding the vertical coherence of the water column.

Finally, we constructed two predictors with the same architecture but different lead-time targets: an FM1 predictor for 1-day forecasting and an FM5 predictor for 5-day forecasting. During inference, FM1 was used to step through the early forecast period so that rapid local structural changes were well resolved, after which FM5 was used to extend the forecast more efficiently. This dual-predictor strategy balances short-range local accuracy and longer-range forecast stability.

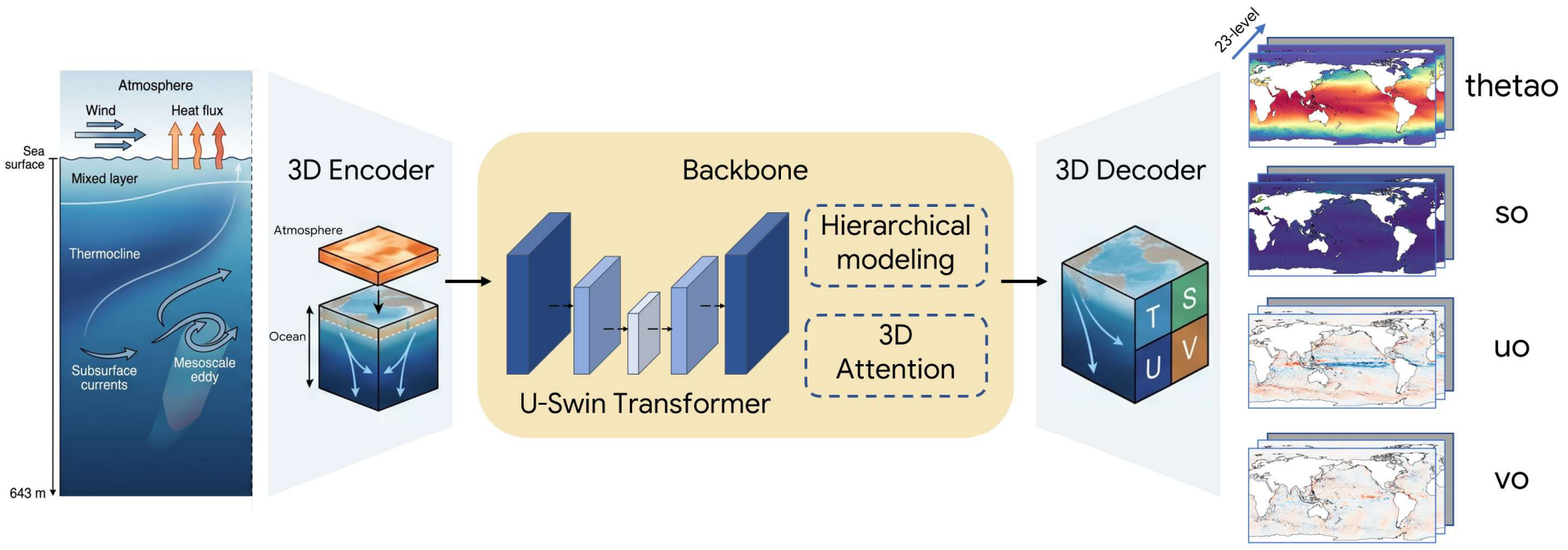

**Fig.6 Model AxiomOcean framework and dual-propagator forecast design.** Architecture of AxiomOcean. A three-dimensional encoder maps the coupled atmosphere-ocean state into a latent representation that retains upper-ocean vertical structure, which is evolved by a U-Swin Transformer backbone with hierarchical modeling and three-dimensional attention and decoded into forecasts of potential temperature (thetao), salinity (so), zonal velocity (uo), and meridional velocity (vo) on 23 vertical levels.

## Loss function and coastal-boundary treatment

Training minimizes the error in predicted state increments. Because a latitude-longitude grid represents unequal physical area across latitudes, we applied latitude-based area weighting in the loss function. Losses were computed only over valid ocean grid cells to avoid contamination from land points.

$$\mathcal{L} = \frac{1}{\sum_{v,k,i,j} w_i \, M_{k,i,j}} \sum_{v=1}^{V} \sum_{k=1}^{D} \sum_{i=1}^{H} \sum_{j=1}^{W} w_i \, M_{k,i,j} \left( \Delta\hat{O}_{v,k,i,j} - \Delta O_{v,k,i,j} \right)^2 \qquad (2)$$

Although supervision was restricted to ocean points, we did not apply an additional land-sea mask inside the attention calculations. This choice preserves near-coastal geometric context and avoids prematurely severing spatial relationships at boundaries, which can otherwise reduce the model's ability to infer coherent structure near coasts and marginal seas.

## Training strategy

Global three-dimensional forecasting at 1/12° resolution is memory-intensive. To enable training at full resolution while preserving three-dimensional structure, we trained the model on eight NVIDIA A100 GPUs (80 GB each) for approximately four weeks, using distributed training together with memory-saving strategies including pipeline parallelism and gradient checkpointing.

We further adopted a two-stage training strategy. In the first stage, the model was trained on shorter temporal windows so that it first learned basic dynamical and thermodynamical adjustment relationships. In the second stage, training continued on the full temporal span to expose the model to broader seasonal backgrounds and regional regimes, thereby improving generalization.

## Data Availability

GLORYS12 and ERA5 are publicly available through the Copernicus Marine Service and the Copernicus Climate Data Store.

## Code Availability

The source code will be made publicly available upon publication.

## Competing Interests

The authors declare no competing interests.